%% file: ifacconf.tex
\documentclass{ifacconf}

\usepackage{graphicx}      
\usepackage{natbib}        
\usepackage{amssymb}
\usepackage{amsmath}
\usepackage{xcolor}
\usepackage{subcaption}
\usepackage{booktabs}
\usepackage{adjustbox}
\newcommand\norm[1]{\left\lVert#1\right\rVert}

\pdfoutput=1
\begin{document}
\begin{frontmatter}

\title{A Distance-Geometric Method for Recovering Robot Joint Angles From an RGB Image\thanksref{footnoteinfo}} 

\thanks[footnoteinfo]{This work has been supported by the European Regional Development Fund under the grant KK.01.1.1.01.0009 (DATACROSS).}
\thanks[footnoteinfo]{© 2023 the authors. This work has been accepted to IFAC for publication under a Creative Commons Licence CC-BY-NC-ND}

\author[First]{Ivan Bilić} 
\author[]{{Filip Marić}\textsuperscript{\normalfont{*, **}}} 
\author[First]{Ivan Marković}
\author[First]{Ivan Petrović}

\address[First]{University of Zagreb, Faculty of Electrical Engineering and Computing, Laboratory for Autonomous Systems and Mobile Robotics (e-mail: name.surname@fer.hr).}
\address[Second]{University of Toronto, Space and Terrestrial Autonomous Robotic Systems Laboratory (e-mail: name.surname@mail.utoronto.ca)}

\begin{abstract}
Autonomous manipulation systems operating in domains where human intervention is difficult or impossible (e.g., underwater, extraterrestrial or hazardous environments) require a high degree of robustness to sensing and communication failures.
Crucially, motion planning and control algorithms require a stream of accurate joint angle data provided by joint encoders, the failure of which may result in an unrecoverable loss of functionality.
In this paper, we present a novel method for retrieving the joint angles of a robot manipulator using only a single RGB image of its current configuration, opening up an avenue for recovering system functionality when conventional proprioceptive sensing is unavailable.
Our approach, based on a distance-geometric representation of the configuration space, exploits the knowledge of a robot's kinematic model with the goal of training a shallow neural network that performs a 2D-to-3D regression of distances associated with detected structural keypoints.
It is shown that the resulting Euclidean distance matrix uniquely corresponds to the observed configuration, where joint angles can be recovered via multidimensional scaling and a simple inverse kinematics procedure.
We evaluate the performance of our approach on real RGB images of a Franka Emika Panda manipulator, showing that the proposed method is efficient and exhibits solid generalization ability.
Furthermore, we show that our method can be easily combined with a dense refinement technique to obtain superior results. 
\end{abstract}


\end{frontmatter}

\section{Introduction}
Autonomous manipulation systems are ideal for performing various tasks in environments where human presence is limited, such as underwater or orbital laboratories, as well as hazardous (e.g., radioactive, toxic) environments.
In addition to effective planning and control algorithms, these systems require a high degree of robustness to sensing and communication failures, as a timely intervention by humans may be impossible.
We propose a method for recovering the joint angles (i.e., the configuration) of an articulated robotic manipulator using only a single RGB image, providing an alternative source of proprioceptive data that can be used when data from joint encoders is unavailable.
This is a challenging task for multiple reasons: fundamentally different robot configurations may result in similar images and certain configurations may have diminished observability due to physical occlusion by other parts of the robot.
However, even a rough estimate of the joint configuration may enable the use of simple control methods to steer the robot to an approximate desired state, enabling the planning and execution of critical recovery protocols (\cite{vision_based_control}).
The problem of recovering a robot configuration from spatial constraints such as gripper pose is known as inverse kinematics and features a variety of well known solutions (\cite{lynch2017modern}).
However, previous work has also explored instances of this problem where spatial constraints may result from a variety of sensing modalities, such as depth or RGB images.
\cite{Widmaier} use synthetic depth images to train semantic classifiers for direct joint angle regression in order to estimate the robot arm pose.
\cite{c7} also use depth images to train a random forest classifier for pixel-wise part classification, while using joint encoder readings to initialize an incremental update scheme. 
%
%
Conversely, our method aims to recover joint configuration using only a single RGB image, which necessitates first finding the appropriate set of spatial constraints using 2D-to-3D regression.

The constraints resulting from regressed 3D keypoints may be under-determined and therefore correspond to multiple configurations.
Instead, we use a distance-geometric model that integrates structural data (e.g., link lengths) to remove ambiguity.
Distance geometry is highly relevant for applications such as molecular conformation, sensor network localization (SNL) and statics (\cite{c13}). 
For instance, SNL is commonly framed as an Euclidean distance matrix (EDM) completion problem (\cite{dokmanic}) and tackled through semidefinite programming (SDP) (\cite{sdp_biswas}).
\cite{TRO_Maric} consider a large class of articulated robot manipulators and elaborate on the equivalence of the distance geometry problem and distance-based inverse kinematics.
Furthermore, \cite{EDM_human_pose_estimation} tackles the problem of 3D human pose estimation from a single RGB image and demonstrates that representing human poses with EDMs instead of Cartesian coordinates results in more precise and less ambiguous pose estimates. Our work is partly inspired by these observations, and we show that the problem of recovering robot’s joint angles is in general highly related to distance geometry.

To the best of our knowledge, the only similar approaches to ours are that of \cite{craves} and \cite{c5}, in the sense that only a single RGB image is used as an input for joint angle estimation.
\cite{craves} train a joint keypoint detector to recognize a predefined, specifically chosen set of 17 keypoints in the image displaying a 4-DoF toy robot. 
These keypoints are then fed to a nonlinear non-convex 2D-to-3D optimization algorithm in order to recover the 6D pose together with joint angles. 
Although our distance-geometric method is also keypoint-based, it only requires a number of keypoints equal to the robot's DoF, placed in a robot-invariant manner. 
Furthermore, the optimization proposed in \cite{craves} is complex, which diminishes its potential to improve from scaling the model and data size. 
On the other hand, \cite{c5} use a dense, rendering-based deep iterative matching approach (\cite{li2018deepim}) to jointly learn the 6D camera-to-robot pose and joint angle updates. 
Although primarily concerned with estimating the 6D pose, \cite{c5} also demonstrate that joint angles, if unknown, can be accurately reconstructed, at least with a sufficient number of iterations. 
Overall, the dense approach seems to generally outperform sparse approaches in terms of accuracy due to incorporating global information, at the cost of high computation time.
We demonstrate that using our method in conjunction with a dense refinement offers the best of both worlds.

In this paper, we propose a novel method for recovering the joint angles of an articulated robotic manipulator using only a single RGB image, based on a distance-geometric representation of the configuration space and the knowledge of a robot’s kinematic model\footnote{https://github.com/iwhitey/distance-geometric-robot-joint-angles}.
Instead of training a single large model to directly predict the solution, our method divides the problem into a set of smaller sub-problems in a theoretically justified manner.
First, state-of-the-art keypoint detectors (\cite{c17}) are used to detect joint keypoints in the image corresponding to the robot's joints, which are insufficient for configuration recovery on their own.
Then, our method takes the full set of inter-point distances and uses a learned 2D-to-3D regression to produce an EDM corresponding to the associated 3D keypoints.
Following the approach in (\cite{TRO_Maric}) this EDM is extended to include distances between auxiliary points determined by the robot's structure, removing ambiguity related to joint angle recovery. 
%
%
Given a complete EDM, joint angles can be computed using parameter-free transformations; classical multidimensional scaling (MDS) and kinematic transformations.
The former maps an EDM to a geometrically centered set of 3D points, while the latter calculates the joint angles based on these points, forming a fully-differentiable set of kinematic transformations\footnote{We refer to this set of transformations as an IK layer in the rest of the paper.} that supports batching and can be ran on a GPU.
In addition to generating the complete EDM, ground-truth joint angles are used to compute the loss in the configuration space.
Finally, our method is evaluated on a large set of real images displaying a 7-DoF robot arm in various configurations. 
We opt for a shallow architecture for all our experiments since the primary goal of this work is to develop a geometrically meaningful learning-based framework.
The proposed method exhibits solid generalization ability, while being simple and computationally efficient.

\section{Methodological background}\label{EDMs section}

Euclidean distance geometry is an important tool in several applications whose aim is to reconstruct a complete set of distances (or points that realize them) in Euclidean space, given an incomplete set of distances as an input.
In addition to a small subset of distance geometry which relates to EDMs, this section describes a kinematics procedure responsible for generating a set of 3D points sufficient for recovering the robot's configuration as well as distance constraints emerging from kinematics.

\subsection{Euclidean distance matrices}\label{EDMs}
Let $\mathbf{P} \in \mathbb{R}^{n \times d}$ denote a matrix representing a set of $n$ points in a $d$-dimensional Euclidean space.
Then, the pairwise distances $d_{u, v}$ between points can be calculated using the Euclidean norm:
\begin{equation}\label{eq:euclidean_norm}
\mathit{\quad d_{u, v} = \norm{\mathbf{p}_{u} - \mathbf{p}_{v}}}.
\end{equation}

For the sake of notation simplicity, EDMs and individual distances are assumed to be squared in the remaining of the paper. Expansion of (\ref{eq:euclidean_norm}) reveals that EDM is a function of the Gram matrix $\mathbf{G}=\mathbf{P}\mathbf{P}^{\top}$:

\begin{equation}\label{eq:distance from gram matrix}
\operatorname{edm}(\mathbf{G})=\operatorname{diag}(\mathbf{G}) \mathbf{1}^{\top} + \mathbf{1} \operatorname{diag}(\mathbf{G})^{\top}-2 \mathbf{G},
\end{equation}

with $\operatorname{diag}(\mathbf{G})$ representing the diagonal entries of $\mathbf{G}$ in the form of a column vector and $\mathbf{1}$ stands for column vector filled with ones. 

Equation (\ref{eq:distance from gram matrix}) establishes a one-way connection between an EDM and a Gram matrix. Consider an inverse problem, i.e. recovering the set of points that generated the distance matrix $\mathbf{D}$. Let $\mathbf{D}$ be a squared EDM. Then, a Gram matrix that satisfies (\ref{eq:distance from gram matrix}) can be determined via
\begin{equation}\label{eq:gram matrix from edm}
\mathbf{G} = -\frac{1}{2}\mathbf{J}\mathbf{D}\mathbf{J},
\end{equation}
where
\begin{equation}\label{eq:geometric centering matrix}
\mathbf{J} = \mathbf{I} - \frac{1}{N}\mathbf{1}\mathbf{1^{T}}
\end{equation}

denotes a geometric centering matrix. Moreover, $\mathbf{G}$ is a real symmetric matrix, hence it can be factored into a canonical form via eigenvalue decomposition:

\begin{equation}\label{eq:evd}
\mathbf{G} = \mathbf{U}\mathbf{\Lambda}\mathbf{U^{T}}
\end{equation}

where $\mathbf{\Lambda} = \operatorname{diag}\left(\lambda_0, \lambda_1, \ldots, \lambda_{n-1}\right)$ contains the non-negative eigenvalues $\lambda_{i}$ and U is an orthonormal matrix. Now, assuming the eigenvalues are sorted in the descending order, the point set $\hat{\mathbf{P}} \in \mathbb{R}^{n \times d}$ can finally be recovered by taking:

\begin{equation}\label{eq: point-set from evd}
\hat{\mathbf{P}}^{\top}=\left[\operatorname{diag}\left(\sqrt{\lambda_0}, \sqrt{\lambda_1}, \ldots, \sqrt{\lambda_{d-1}}\right), \mathbf{0}_{d \times N-d}\right] \mathbf{U}^{\top}.
\end{equation}

Computing the collection of points $\hat{\mathbf{P}}$ from a distance matrix $\mathbf{D}$ using (\ref{eq:gram matrix from edm}), (\ref{eq:geometric centering matrix}), (\ref{eq:evd}) and (\ref{eq: point-set from evd}) is also known as classical multidimensional scaling (cMDS). Note how in (\ref{eq: point-set from evd}), all but the $d$ largest eigenvalues are discarded. Assuming that $\mathbf{G}$ is generated by a $d$-dimensional set of points, all but the $d$ largest eigenvalues will be zeros. If this is not the case, we can assume the presence of noise which is handled by the truncation (\cite{dokmanic}).
%
%
Additionally, plugging the estimated $\hat{\mathbf{P}}$ in (\ref{eq:distance from gram matrix}) yields an EDM that equals $\mathbf{D}$. However, distances are preserved under rigid transformations, thus $\hat{\mathbf{P}}$ and the original $\mathbf{P}$ are not generally equal. Absolute position and orientation of the point set can be recovered through Procrustes analysis (\cite{procrustes}), assuming that a set of at least $d$ fixed points (i.e. anchors) is known beforehand. Then, a rigid transformation that aligns the anchors in $\mathbf{P}$ to their corresponding points in $\hat{\mathbf{P}}$ can be found. Finally, the original set of points $\mathbf{P}$ can be recovered by applying the obtained rigid transformation to all the points in $\hat{\mathbf{P}}$.





\subsection{Distance-based kinematics}\label{auxiliary points}
\begin{figure}[htbp]
\centering
\includegraphics[width=0.3\textwidth]{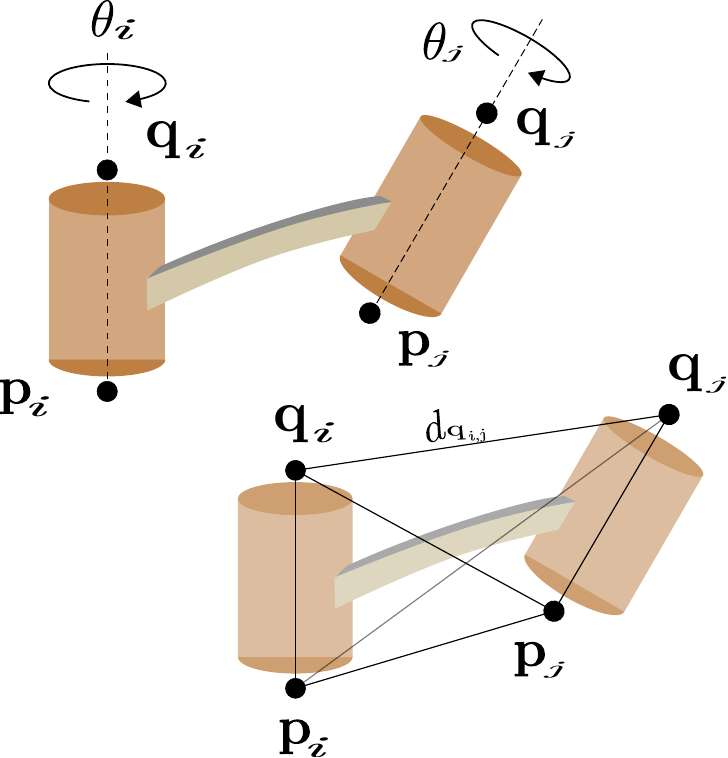}
\caption{Visualization of neighbouring revolute joints. Each joint is rotated by an angle $\theta$ around its rotation axis (dashed lines). Auxiliary points $\mathbf{q}$ are placed along these axis. Points $\mathbf{p}$ correspond to the position of coordinate frames defined in the center of their respective joints (placed below for better visibility).}
\label{fig:dgp}
\end{figure}

Consider an $n$-DoF robotic manipulator comprised of single-axis revolute joints, forming a kinematic chain.
The procedure introduced in~\cite{TRO_Maric} defines a sparse set of points whose positions are sufficient for recovering the full set of joint angles determining the robot's configuration.
As shown in Figure \ref{fig:dgp}, a set of points $\mathbf{p}_{i}$ centered at the joint coordinate frames are introduced, which we associate with the keypoints detected in the RGB image of the robot by our network.
Then, "virtual" auxiliary points $\mathbf{q}_i$ are placed at a unit distance along the joints' rotation axes $\tilde{\mathbf{z}}$ using joints' orientation $\mathbf{R}_i$
\begin{equation}\label{eq:q points from p points}
\mathbf{q}_i = \mathbf{p}_i + \mathbf{R}_i\tilde{\mathbf{z}} ,
\end{equation}
adding information on the relative orientation of neighbouring joints required for joint angle recovery.
Finally, the model is completed with the addition of three points, $\mathbf{x}$, $\mathbf{y}$ and $\mathbf{z}$, corresponding to the root coordinate frame, defined by distances
\begin{equation}\label{eq:base frame distances}
\begin{aligned}
\norm{\mathbf{p}_{0} - \mathbf{x}} &= \norm{\mathbf{p}_{0} - \mathbf{y}}= \norm{\mathbf{p}_{0} - \mathbf{z}} = 1, \\
\norm{\mathbf{x} - \mathbf{y}} &= \norm{\mathbf{x} - \mathbf{z}} = \norm{\mathbf{y} - \mathbf{z}} = \sqrt{2}.
\end{aligned}
\end{equation}
The proof in~\cite{TRO_Maric} shows that the distances between these points are sufficient for recovering the full set of joint angles for a large variety of manipulator structures.

%
Our training data is generated from ground truth joint angle vectors $\mathbf{\Theta} \in \mathcal{C}$ corresponding to the robot observed in the image $\mathcal{I}$, where $\mathcal{C} \subseteq \mathbb{R}^n$ represents the configuration space. This allows us to simply generate the full distance-geometric robot description. 
Each data point is constructed by sequentially taking the position $\mathbf{p}_i$ and orientation $\mathbf{R}_i$ of a parent joint $i$, as well as its joint angle $\theta_{i}$, giving the position and orientation of the child joint $j$ as
\begin{equation}\label{eq:neighbour position and orientation}
\begin{aligned}
\mathbf{R}_j &=\mathbf{R}_i \mathbf{R}_z\left(\theta_{i}\right) \mathbf{R}_{i, j}, \\ 
\mathbf{p}_j &=\mathbf{p}_i + \mathbf{R}_i \mathbf{R}_z\left(\theta_i\right) \mathbf{p}_{i, j}.
\end{aligned}
\end{equation}
The relative positions and orientations $\mathbf{p}_{i, j}$ and $\mathbf{R}_{i, j}$ of neighbouring joints, are completely defined by the robot's structure.
Thus, the distances between the set of the four neighbouring points can also be defined from structural parameters as
\begin{equation}\label{eq:fixed neighbouring distances}
\begin{aligned}
\norm{\mathbf{p}_i - \mathbf{q}_i} &= \norm{\mathbf{p}_j - \mathbf{q}_j} = 1 \\
\norm{\mathbf{p}_i - \mathbf{p}_j} &= \norm{\mathbf{p}_{i,j}} \\
\norm{\mathbf{p}_i - \mathbf{q}_j} &= \norm{\mathbf{p}_{i,j} + \mathbf{R}_{i, j}\tilde{\mathbf{z}}} \\
\norm{\mathbf{q}_i - \mathbf{p}_j} &= \norm{\mathbf{p}_{i,j} - \tilde{\mathbf{z}}} \\
\norm{\mathbf{q}_i - \mathbf{q}_j} &= \norm{\mathbf{p}_{i,j} - \tilde{\mathbf{z}} + \mathbf{R}_{i,j}\tilde{\mathbf{z}}},
\end{aligned}
\end{equation}
indicating their invariance to changes in joint angles.
Conversely, the distances corresponding to the remaining non-neighbouring pairs exist as a function of the configuration $\mathbf{\Theta}$. 
Using the constructive procedure in (\ref{eq:q points from p points}) and (\ref{eq:neighbour position and orientation}), together with distance constraints stated in (\ref{eq:base frame distances}) and (\ref{eq:fixed neighbouring distances}), we construct a complete EDM $\mathbf{D} \in \mathbb{R}^{\hat{n} \times \hat{n}}$ that uniquely (up to a rigid transformation) represents a set of 3D points associated with the configuration $\mathbf{\Theta}$ observed in the RGB image.
%
%
Concretely, for an $n$-DoF articulated robot comprised of single-axis revolute joints - one auxiliary point per joint to determine the axis of rotation, and two additional points are required to fully define the base coordinate frame. Finally, a single data sample used for training the model can thus be formalized as $\mathcal{D} := (\mathbf{\Theta},\mathbf{D})$.

\begin{figure*}[t]
    \centering    
    \def\svgwidth{\textwidth}
    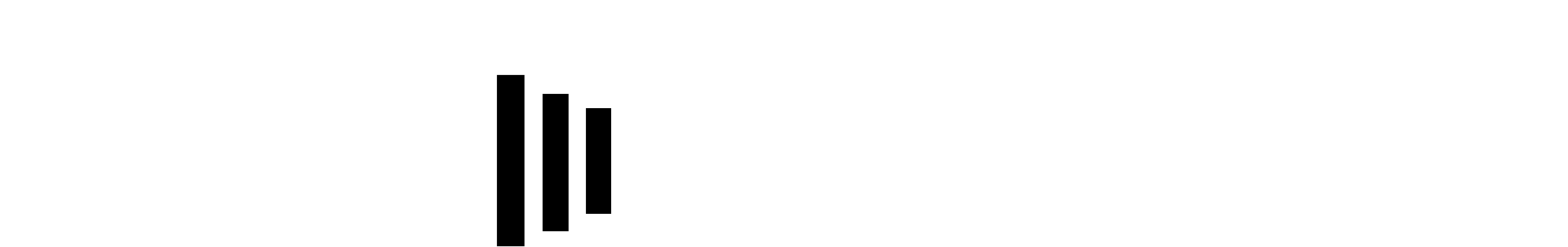
    \caption{System overview. In the image space, EDM is constructed from detected 2D joint keypoints and fed to a 2D-to-3D EDM regression network (black), which outputs the complete EDM in the 3D space. The cMDS layer maps the complete EDM to a set of geometrically centered points, from which an IK layer computes the joint angles. Estimated joint angles are shown as an image of the robot rendered in the respective configuration.}
    \label{fig:system}
\end{figure*}

\section{Method}\label{sec:the_method}
\begin{figure}[b]
\centering
\includegraphics[width=0.4\textwidth]{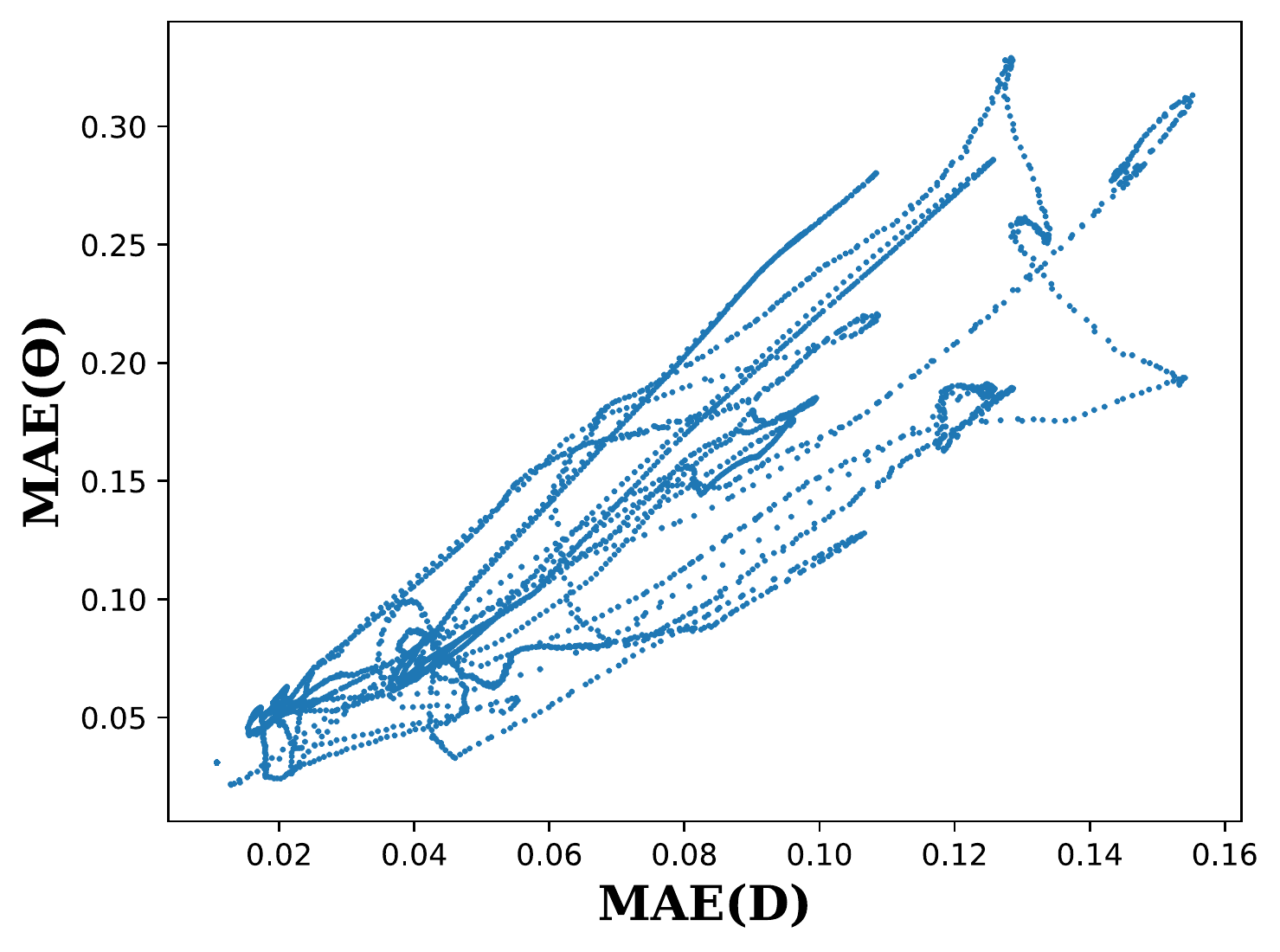}
\caption{Joint angles vs EDMs, using mean absolute error (MAE). The model is trained for EDM regression only. The errors are highly correlated; Pearson correlation coefficient is 0.94.}
\label{fig:correlation}
\end{figure}

This section is dedicated to describing the system displayed in Figure \ref{fig:system} as a pipeline through which the proposed method is implemented. First, given the input RGB image and 2D joint keypoint detections, one needs to bridge the gap between 2D and 3D information.
Theoretically, infinitely many sets of $\hat{n}$ points which are equal up to a rigid transformation, and generated as described in Section \ref{auxiliary points}, can map to a single feasible robot configuration $\mathbf{\Theta}$.
As inter-point distances remain identical irrespective of such point set transformation, we associate a configuration $\mathbf{\Theta}$ with a single unique EDM.
Therefore, instead of directly predicting the points, we predict the corresponding distance matrix. 
We frame the distance matrix regression problem as learning a mapping $\zeta: \mathbb{R}_+^{n \times n} \mapsto \mathbb{R}_+^{\hat{n} \times \hat{n}}$, where $\mathbb{R}^{+} = \{x \in \mathbb{R}: x \geq 0\}$, $n$ equals the number of joints, and $\hat{n}=2n+2$. Note that we don't use auxiliary points in the image space, only keypoints directly corresponding to the robot's joints.
The mapping is implemented as a shallow feedforward neural network that takes an EDM computed from 2D joint keypoints as input and outputs a complete EDM which is treated as if it was generated by a set of 3D points comprised of joint correspondences and auxiliary points.
%
We define a distance-based loss as:
\begin{equation}\label{eq:distance loss function}
L_d = \| \hat{\mathbf{D}} - \mathbf{D}\|_F,
\end{equation}

where $\|\cdot\|_F$ denotes Frobenius norm, while $\hat{\mathbf{D}}$ and $\mathbf{D}$ are the predicted and ground-truth EDMs, respectively. 
Due to the fact that EDMs are symmetric matrices, the actual implementation works with upper-triangular elements from which the full EDM is computed afterwards. 
This amounts to using $n(n-1)/2$ elements instead of $n^2$ with no loss of information. 
The ground-truth EDMs are computed from ground-truth configurations $\mathbf{\Theta}$ via a function $f: \mathbf{\Theta}^n \mapsto \mathbf{D}^{\hat{n} \times \hat{n}}$ based on equations described in Section \ref{auxiliary points}. 
Figure \ref{fig:correlation} demonstrates that EDM regression and joint angle recovery are highly related, by depicting a mean absolute joint angle error as a function of mean absolute EDM error, using the Kinect dataset (unseen during the learning process). Note that the model used for this figure is trained for EDM regression, i.e. using the loss function (\ref{eq:distance loss function}), while the joint angles are only computed during inference via cMDS and IK layers.
We use our library\footnote{https://github.com/utiasSTARS/graphIK} for most of the distance-geometry and kinematics-related computation.

After the complete EDM $\mathbf{D} \in \mathbb{R}^{\hat{n} \times \hat{n}}$ is estimated, the cMDS layer (a set of fixed, differentiable transformations described in Section \ref{EDMs section}) is used to obtain the geometrically centered set of points which generate the respective EDM. This mapping can be formally defined as $\Omega : \mathbb{R}_+^{\hat{n} \times \hat{n}} \mapsto \mathbf{P}^{\hat{n} \times d}$. 
The set of points is then fed to an IK layer which computes the joint angles $\mathbf{\Theta}$. Note that what we refer to as an IK layer is not an inverse kinematics solver; it is a sequence of differentiable kinematics transformations that, given the estimated set of points, compute the joint angles by computing the respective coordinate frame positions and orientations together with the axis of rotation for each joint (\cite{TRO_Maric}).
The default configuration (corresponding to zero joint angles) of the robot is known from its model (an Unified Robotics Description Format file), hence the joint angles can be computed. This allows us to define a loss in the configuration space between the predicted and ground-truth configuration:
\begin{equation}\label{eq:joint angle loss function}
L_c = \lvert\hat{\mathbf{\Theta}} - \mathbf{\Theta}\rvert,
\end{equation}
with $\lvert\cdot\rvert$ denoting the L1 norm. We use a linear combination of the two losses as a final loss to train the model:
\begin{equation}\label{eq:final loss}
L = L_c + \lambda L_d, 
\end{equation}
with $\lambda$ set to $0.5$. 
Note that $L_c$ causes the gradients to be propagated through IK and cMDS layers, while $L_d$ is applied directly at the output of the EDM regression network and serves to provide the model with additional information.

The EDM regression network is comprised of two $Dense \rightarrow BatchNorm \rightarrow ReLU \rightarrow Dropout$ layers, and an output $Dense \rightarrow ReLU$ layer. The respective hidden and output layer sizes are $512 \rightarrow 512 \rightarrow 120$, which amounts to $\sim$ 350k parameters. The output size is determined by\footnote{The number of non-zero elements in strictly upper triangular EDM.} $\hat{n}(\hat{n}-1)/2$, where $\hat{n}=2n+2$ and $n=7$ for the Franka Emika Panda robot. Among other necessary conditons, EDM elements must be positive, which is enforced by the last ReLU (\cite{hinton2010rectified}) activation. We use Adam (\cite{c14}) for optimization, with initial learning rate $\alpha=1e-3$, linear warmup (\cite{c16}) over 2k iterations, a batch size of $64$ EDMs, and a dropout (\cite{srivastava2014dropout}) ratio of $0.5$. The training is carried out for $100$ epochs and learning rate is reduced by a factor of $2$ after $50$ epochs. The model is initialized as proposed by \cite{c15}.


\section{Experiments}

All our experiments were conducted on images of a 7-DoF Franka Emika Panda robot observed in various configurations, using three different datasets. We report mean absolute error as a joint angle error metric.
Training and evaluation were carried out on a single NVIDIA RTX A5000. Using the setup described in Section \ref{sec:the_method}, inference takes $1.6$ms and requires 1.5GB of GPU memory. In reality, it would be limited by the running time of the chosen keypoint detector.
%
%

\subsection{Dataset}
For our experiments, we use the DREAM dataset introduced by \cite{c17} in their recently proposed state-of-the-art method for single-view camera-to-robot pose estimation. 
The dataset is comprised of real and synthetic parts. 
The synthetic part is photorealistic and generated with domain randomization. 
We focus on the real part of the dataset, which is split into $4$ different Panda-3CAM datasets - Realsense, Azure, Kinect, and Orb which contain 5944, 6394, 4966 and 32315 samples, respectively. 
Each of these datasets is comprised of RGB images of the 7-DoF Franka Emika Panda robotic manipulator, captured using different cameras with different intrinsic parameters. 
The robot is observed in various configurations, including images with joint occlusions and even out-of-view joints. Besides RGB images, the datasets contain 2D joint keypoint annotations together with their 3D correspondences and ground-truth robot configuration. 
The camera-to-robot pose is different and fixed for each dataset, except for Orb which is captured from 27 different viewpoints. For all our experiments we used 8x subsampled version of the Orb dataset for training, which we refer to as Orb in the rest of the section. We automatically adjusted all the 2D joint keypoint annotations in all the datasets so that they match the Panda's coordinate frame definitions exactly.

\subsection{Results}
%
The results of applying our method on the Panda-3CAM datasets are displayed in Table \ref{table:results_our_method}. 
For evaluating on Kinect and Azure datasets the model is trained on Realsense and Orb, while for Realsense evaluation we trained it on the Kinect and Orb datasets. 
The results show that our method gives solid joint angle approximations on unseen data, while being simple and computationally efficient. 
The top 50\% predictions mostly correspond to images which display configurations relatively close to coplanar with respect to the image plane, for a given robot-camera pose. 
However, the datasets also contain images with joint occlusions and highly non-coplanar configurations with respect to the image plane, making the task more difficult for a sparse, keypoint-based method such as ours. By manual inspection, we detected that Azure dataset contains more such images compared to the other two datasets, which reflects on the results. The more thorough error analysis is left for future work.

\begin{table}[!t] 
\centering
\captionsetup{width=0.8\columnwidth}
\caption{Results on the Panda-3CAM datasets. The mean absolute joint angle error and standard deviation are reported. The error corresponds to predictions on all test images, or top 50\% predictions closest to the ground-truth joint angles.}
\begin{tabular}{cccc}\toprule
Dataset & num. images & all [$10^{\circ}$] & top $50\%$ [$10^{\circ}$]\\\midrule
Realsense & 5944 & 1.261 $\pm$ 0.218 & 0.562 $\pm$ 0.08\\
Kinect & 4966 & 1.061 $\pm$ 0.2 & 0.344 $\pm$ 0.04\\
Azure & 6394 & 1.433 $\pm$ 0.264 & 0.733 $\pm$ 0.092\\\bottomrule
\end{tabular}
\label{table:results_our_method}
\end{table}

\begin{figure}[htbp]
\vspace{0.1cm}
\centering
    \begin{subfigure}[t]{0.15\textwidth}
        \centering
        \includegraphics[width=\textwidth]{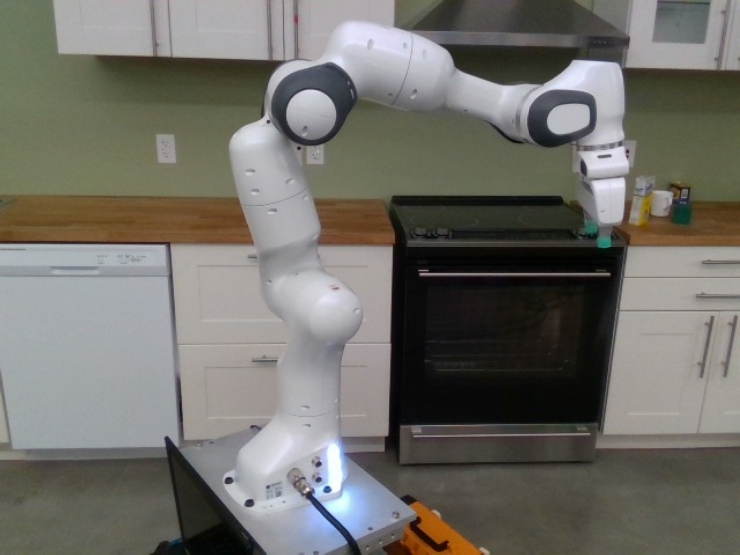}
    \end{subfigure}
    \hfill
    \begin{subfigure}[t]{0.15\textwidth}
        \centering
        \includegraphics[width=\textwidth]{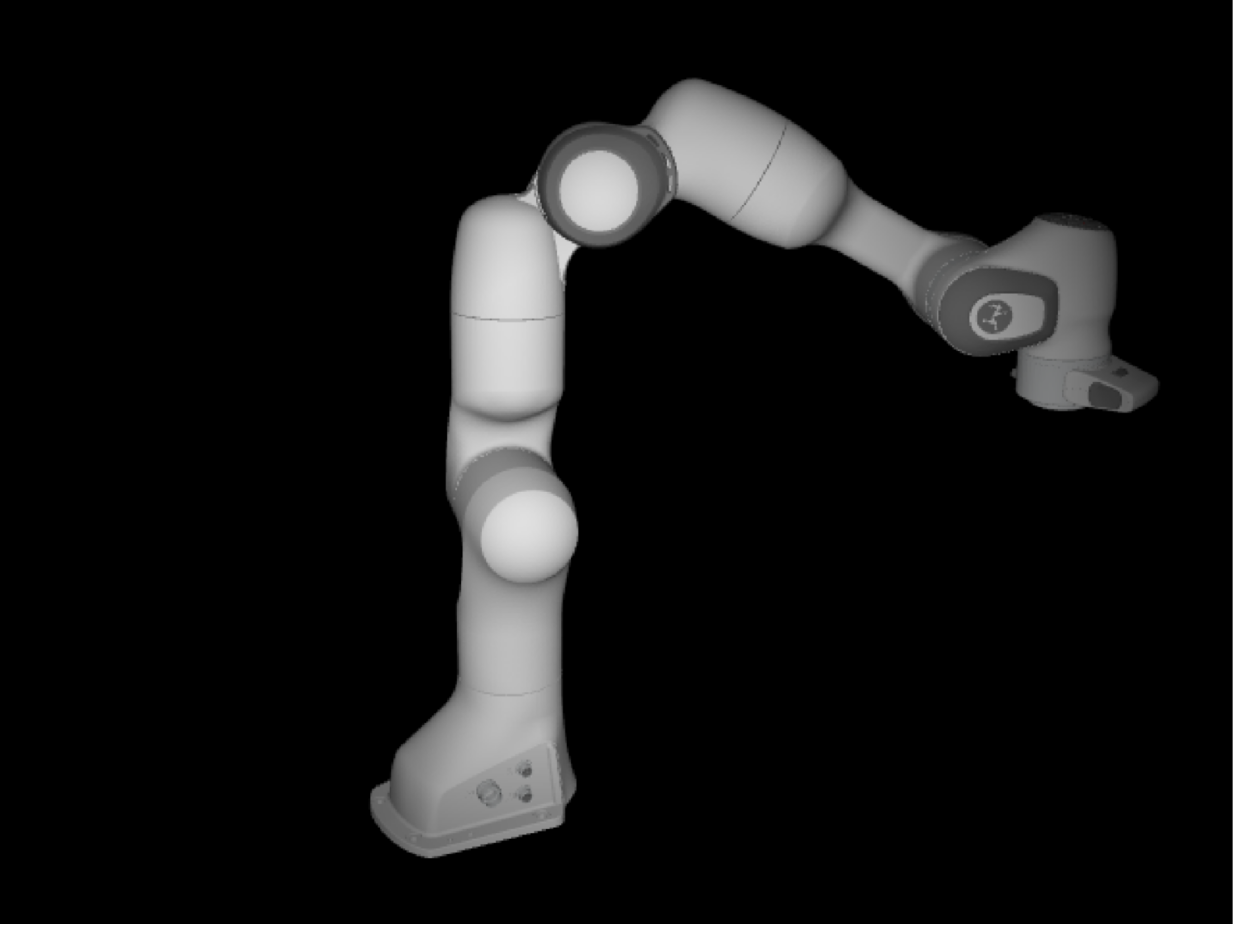}
    \end{subfigure}
    \hfill
    \begin{subfigure}[t]{0.15\textwidth}
        \centering
        \includegraphics[width=\textwidth]{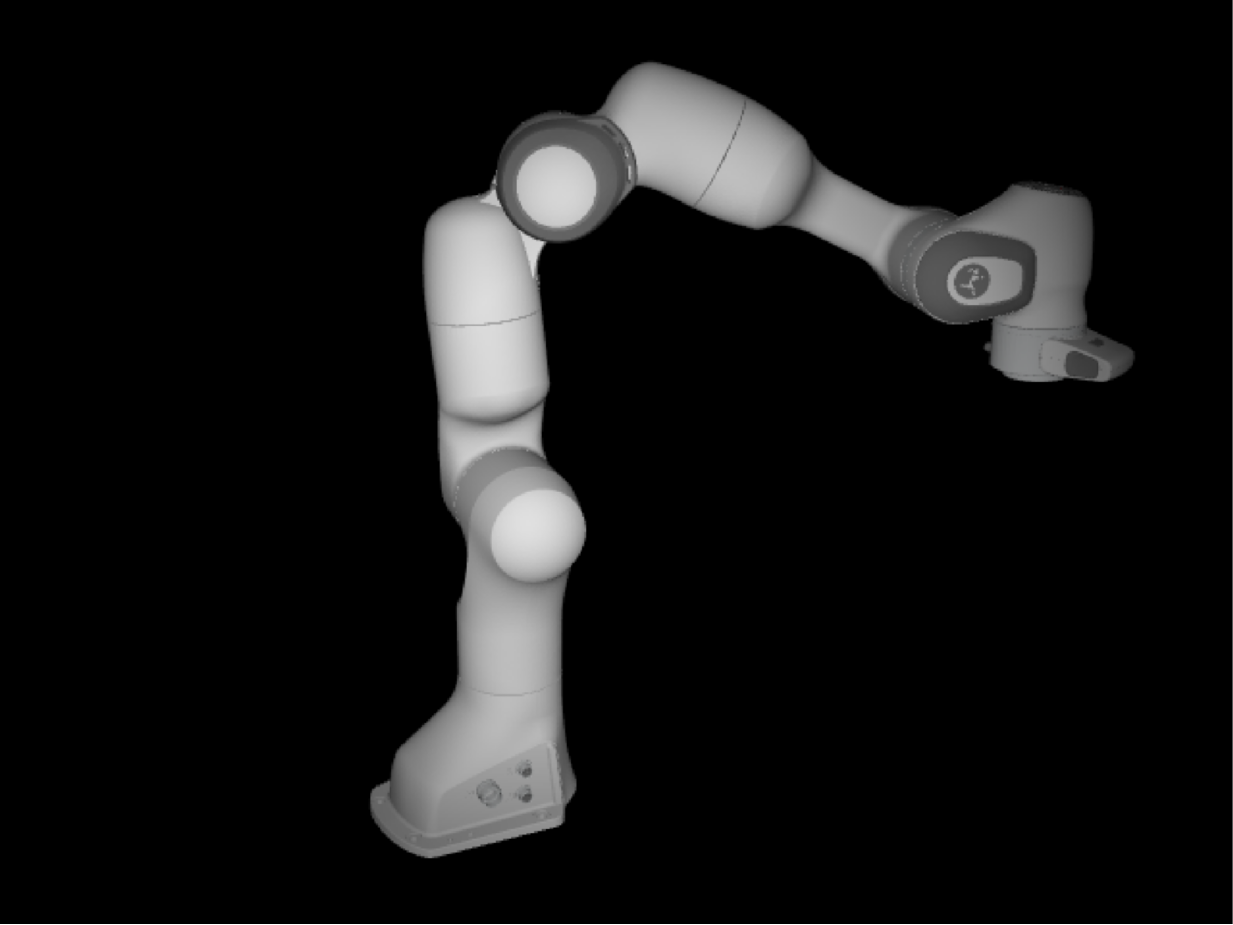}
    \end{subfigure}
        \caption{Input image (left) and rendered joint angle predictions -  our method (middle), our method with refinement (right)}
        \label{fig:render_ours_and_robopose}
\end{figure}

\subsection{Deep refinement}
The proposed distance-geometric method exhibits solid generalization in addition to being lightweight, thus it can be easily used in conjunction with a refinement procedure without introducing noticeable computational complexity. 
%
%
To this end, we use the publicly available pretrained RoboPose model, introduced by \cite{c5} and trained using a deep iterative matching procedure (\cite{li2018deepim}) on the DREAM dataset.
This procedure can be briefly described as follows.
First, the joint angles are initialized randomly within joint angle limits and used to render an RGB image of the 3D robot model in this configuration. 
Then, the rendered and input RGB images are fed to a convolutional backbone. 
The backbone outputs the 6D pose update together with a joint angle residual which are used to update the input and the process is repeated iteratively. 
%


We combine our method with RoboPose by using it to initialize the deep refinement procedure. The results are shown in Table \ref{table:results_refinement}.
Clearly, the combined approach outperforms both our method (Table \ref{table:results_our_method}) and RoboPose applied independently. This is because our model provides a good initial guess, thereby making the refinement task much easier in contrast to using a feasible random configuration as an initial guess. 
An exemplary robot configuration estimated by our method and the combined approach is displayed in Figure \ref{fig:render_ours_and_robopose} in the form of a rendered image.
Furthermore, the goal is to obtain an accurate estimation using as few iterations as possible, hence the results were generated using 3 iterations, where each iteration requires a rendering operation and a forward pass of a deep CNN backbone.
On the contrary, RoboPose, when applied independently, requires at least 10 iterations to achieve similar accuracy. Note that we have not trained the model from scratch in order to adjust it to our initialization - it is pretrained using a feasible random configuration. 
Since our method provides an estimate much closer to the actual solution, we would expect retraining the model to introduce further improvements.
%

\begin{table}[htbp] 
\centering
\captionsetup{width=0.9\columnwidth}
\caption{Results on the Panda-3CAM datasets before and after combining the deep refinement model with our method. The mean absolute joint angle error is reported on all images.}
\begin{tabular}{cccc}\toprule
Dataset & \#images & RoboPose [$10^{\circ}$] & Ours + RoboPose [$10^{\circ}$]\\
\midrule
Realsense & 5944 & 1.221 & 0.585\\
Kinect & 4966 & 1.368 & 0.613\\
Azure & 6394 & 0.93 & 0.544\\
\bottomrule
\end{tabular}
\label{table:results_refinement}
\end{table}


\section{Conclusion}
In this paper, we have proposed a novel distance-geometric framework for recovering the joint angles determining the configuration of the robot from a single RGB image.
In addition to being geometrically meaningful, our method is computationally efficient and exhibits solid generalization ability when tested on a large set of images displaying a state-of-the-art 7-DoF robot. 
We also show that, due to its computational efficiency, it can be easily used in conjuction with a dense refinement approach to obtain superior results. 
We believe that a modular approach is promising in the long-term, i.e. tackling the larger problem through a set of smaller, simpler problems.
If one can detect the joint keypoints and recover the respective EDM in the 3D space accurately, the joint angles can accurately be recovered since the leap from EDMs to joint angles is done via deterministic, parameter-free transformations. 
This is valuable since interpreting and analyzing deep models is often hard due to their nature; thus, if possible, it is appealing to design models that aim to solve smaller pieces of the problem. 
Finally, due to the sensitivity of sparse methods, achieving robustness requires incorporating global, dense information, but in a computationally efficient way.
As future work, we intend to explore the adequacy of different architectures for this task, including scaling the data size and model capacity in order to investigate the potential of our approach more thoroughly.
%

\bibliography{ifacconf}             

\end{document}

%% file: plots/system_with_text_small.pdf_tex
\begingroup%
  \makeatletter%
  \providecommand\color[2][]{%
    \errmessage{(Inkscape) Color is used for the text in Inkscape, but the package 'color.sty' is not loaded}%
    \renewcommand\color[2][]{}%
  }%
  \providecommand\transparent[1]{%
    \errmessage{(Inkscape) Transparency is used (non-zero) for the text in Inkscape, but the package 'transparent.sty' is not loaded}%
    \renewcommand\transparent[1]{}%
  }%
  \providecommand\rotatebox[2]{#2}%
  \newcommand*\fsize{\dimexpr\f@size pt\relax}%
  \newcommand*\lineheight[1]{\fontsize{\fsize}{#1\fsize}\selectfont}%
  \ifx\svgwidth\undefined%
    \setlength{\unitlength}{573.1602286bp}%
    \ifx\svgscale\undefined%
      \relax%
    \else%
      \setlength{\unitlength}{\unitlength * \real{\svgscale}}%
    \fi%
  \else%
    \setlength{\unitlength}{\svgwidth}%
  \fi%
  \global\let\svgwidth\undefined%
  \global\let\svgscale\undefined%
  \makeatother%
  \begin{picture}(1,0.15683126)%
    \lineheight{1}%
    \setlength\tabcolsep{0pt}%
    \put(0,0){\includegraphics[width=\unitlength,page=1]{system_with_text_small.pdf}}%
    \put(0.61102616,0.04775238){\color[rgb]{0,0,0}\makebox(0,0)[lt]{\lineheight{1.25}\smash{\begin{tabular}[t]{l}cMDS\end{tabular}}}}%
    \put(0,0){\includegraphics[width=\unitlength,page=2]{system_with_text_small.pdf}}%
    \put(0.77281859,0.0469003){\color[rgb]{0,0,0}\makebox(0,0)[lt]{\lineheight{1.25}\smash{\begin{tabular}[t]{l}IK\end{tabular}}}}%
    \put(0,0){\includegraphics[width=\unitlength,page=3]{system_with_text_small.pdf}}%
    \put(0.18236233,0.14967949){\color[rgb]{0,0,0}\makebox(0,0)[lt]{\lineheight{1.25}\smash{\begin{tabular}[t]{l}\textit{2D EDM}\\\end{tabular}}}}%
    \put(0.45183843,0.14968141){\color[rgb]{0,0,0}\makebox(0,0)[lt]{\lineheight{1.25}\smash{\begin{tabular}[t]{l}\textit{3D EDM}\\\end{tabular}}}}%
    \put(0.28367781,0.14968141){\color[rgb]{0,0,0}\makebox(0,0)[lt]{\lineheight{1.25}\smash{\begin{tabular}[t]{l}\textit{EDM Regression}\\\end{tabular}}}}%
    \put(0,0){\includegraphics[width=\unitlength,page=4]{system_with_text_small.pdf}}%
  \end{picture}%
\endgroup%

%% file: ifacconf.bbl
\begin{thebibliography}{19}
\providecommand{\natexlab}[1]{#1}
\providecommand{\url}[1]{\texttt{#1}}
\providecommand{\urlprefix}{URL }
\expandafter\ifx\csname urlstyle\endcsname\relax
  \providecommand{\doi}[1]{doi:\discretionary{}{}{}#1}\else
  \providecommand{\doi}{doi:\discretionary{}{}{}\begingroup
  \urlstyle{rm}\Url}\fi

\bibitem[{Biswas et~al.(2006)Biswas, Liang, Toh, Ye, and Wang}]{sdp_biswas}
Biswas, P., Liang, T.C., Toh, K.C., Ye, Y., and Wang, T.C. (2006).
\newblock Semidefinite programming approaches for sensor network localization
  with noisy distance measurements.
\newblock \emph{IEEE Transactions on Automation Science and Engineering}, 3(4),
  360--371.

\bibitem[{Bohg et~al.(2014)Bohg, Romero, Herzog, and Schaal}]{c7}
Bohg, J., Romero, J., Herzog, A., and Schaal, S. (2014).
\newblock Robot arm pose estimation through pixel-wise part classification.
\newblock In \emph{2014 IEEE International Conference on Robotics and
  Automation (ICRA)}, 3143--3150.

\bibitem[{Dokmanic et~al.(2015)Dokmanic, Parhizkar, Ranieri, and
  Vetterli}]{dokmanic}
Dokmanic, I., Parhizkar, R., Ranieri, J., and Vetterli, M. (2015).
\newblock Euclidean distance matrices: Essential theory, algorithms, and
  applications.
\newblock \emph{IEEE Signal Processing Magazine}, 32(6), 12--30.

\bibitem[{He et~al.(2015)He, Zhang, Ren, and Sun}]{c15}
He, K., Zhang, X., Ren, S., and Sun, J. (2015).
\newblock Delving deep into rectifiers: Surpassing human-level performance on
  imagenet classification.
\newblock In \emph{2015 IEEE International Conference on Computer Vision
  (ICCV)}, 1026--1034.

\bibitem[{Kingma and Ba(2014)}]{c14}
Kingma, D.P. and Ba, J. (2014).
\newblock Adam: A method for stochastic optimization.
\newblock \emph{arXiv preprint arXiv:1412.6980}.

\bibitem[{Labbé et~al.(2021)Labbé, Carpentier, Aubry, and Sivic}]{c5}
Labbé, Y., Carpentier, J., Aubry, M., and Sivic, J. (2021).
\newblock Single-view robot pose and joint angle estimation via render \&
  compare.
\newblock In \emph{2021 IEEE/CVF Conference on Computer Vision and Pattern
  Recognition (CVPR)}, 1654--1663.

\bibitem[{Lee et~al.(2020)Lee, Tremblay, To, Cheng, Mosier, Kroemer, Fox, and
  Birchfield}]{c17}
Lee, T.E., Tremblay, J., To, T., Cheng, J., Mosier, T., Kroemer, O., Fox, D.,
  and Birchfield, S. (2020).
\newblock Camera-to-robot pose estimation from a single image.
\newblock In \emph{2020 IEEE International Conference on Robotics and
  Automation (ICRA)}, 9426--9432.

\bibitem[{Li et~al.(2018)Li, Wang, Ji, Xiang, and Fox}]{li2018deepim}
Li, Y., Wang, G., Ji, X., Xiang, Y., and Fox, D. (2018).
\newblock Deepim: Deep iterative matching for 6d pose estimation.
\newblock In \emph{Proceedings of the European Conference on Computer Vision
  (ECCV)}, 683--698.

\bibitem[{Liberti et~al.(2014)Liberti, Lavor, Maculan, and Mucherino}]{c13}
Liberti, L., Lavor, C., Maculan, N., and Mucherino, A. (2014).
\newblock Euclidean distance geometry and applications.
\newblock \emph{SIAM review}, 56(1), 3--69.

\bibitem[{Lynch and Park(2017)}]{lynch2017modern}
Lynch, K.M. and Park, F.C. (2017).
\newblock \emph{Modern Robotics}.
\newblock Cambridge University Press.

\bibitem[{Ma and Yarats(2021)}]{c16}
Ma, J. and Yarats, D. (2021).
\newblock On the adequacy of untuned warmup for adaptive optimization.
\newblock In \emph{Proceedings of the AAAI Conference on Artificial
  Intelligence}, volume~35, 8828--8836.

\bibitem[{Marić et~al.(2022)Marić, Giamou, Hall, Khoubyarian, Petrović, and
  J.Kelly}]{TRO_Maric}
Marić, F., Giamou, M., Hall, A., Khoubyarian, S., Petrović, I., and J.Kelly
  (2022).
\newblock Riemannian optimization for distance-geometric inverse kinematics.
\newblock \emph{IEEE Transactions on Robotics}, 38(3), 1703--1722.

\bibitem[{Moreno-Noguer(2017)}]{EDM_human_pose_estimation}
Moreno-Noguer, F. (2017).
\newblock 3d human pose estimation from a single image via distance matrix
  regression.
\newblock In \emph{2017 IEEE Conference on Computer Vision and Pattern
  Recognition (CVPR)}, 1561--1570.

\bibitem[{Nair and Hinton(2010)}]{hinton2010rectified}
Nair, V. and Hinton, G.E. (2010).
\newblock Rectified linear units improve restricted boltzmann machines.
\newblock In \emph{Proceedings of the 27th international conference on machine
  learning (ICML-10)}, 807--814.

\bibitem[{Ortenzi et~al.(2018)Ortenzi, Marturi, Mistry, Kuo, and
  Stolkin}]{vision_based_control}
Ortenzi, V., Marturi, N., Mistry, M., Kuo, J., and Stolkin, R. (2018).
\newblock Vision-based framework to estimate robot configuration and kinematic
  constraints.
\newblock \emph{IEEE/ASME Transactions on Mechatronics}, 23(5), 2402--2412.

\bibitem[{Schönemann(1966)}]{procrustes}
Schönemann, P. (1966).
\newblock A generalized solution of the orthogonal procrustes problem.
\newblock \emph{Psychometrika}, 31(1), 1--10.

\bibitem[{Srivastava et~al.(2014)Srivastava, Hinton, Krizhevsky, Sutskever, and
  Salakhutdinov}]{srivastava2014dropout}
Srivastava, N., Hinton, G., Krizhevsky, A., Sutskever, I., and Salakhutdinov,
  R. (2014).
\newblock Dropout: a simple way to prevent neural networks from overfitting.
\newblock \emph{The journal of machine learning research}, 15(1), 1929--1958.

\bibitem[{Widmaier et~al.(2016)Widmaier, Kappler, Schaal, and Bohg}]{Widmaier}
Widmaier, F., Kappler, D., Schaal, S., and Bohg, J. (2016).
\newblock Robot arm pose estimation by pixel-wise regression of joint angles.
\newblock In \emph{2016 IEEE International Conference on Robotics and
  Automation (ICRA)}, 616--623.

\bibitem[{Zuo et~al.(2019)Zuo, Qiu, Xie, Zhong, Wang, and Yuille}]{craves}
Zuo, Y., Qiu, W., Xie, L., Zhong, F., Wang, Y., and Yuille, A. (2019).
\newblock Craves: Controlling robotic arm with a vision-based economic system.
\newblock In \emph{2019 IEEE/CVF Conference on Computer Vision and Pattern
  Recognition (CVPR)}, 4209--4218.

\end{thebibliography}
